# Towards Successful Collaboration: Design Guidelines for AI-based Services enriching Information Systems in Organisations

**Nicholas R. J. Frick**

Professional Communication in Electronic Media/Social Media
University of Duisburg-Essen
Duisburg, Germany
Email: nicholas.frick@uni-due.de

**Felix Brünker**

Professional Communication in Electronic Media/Social Media
University of Duisburg-Essen
Duisburg, Germany
Email: felix.bruenker@uni-due.de

**Björn Ross**

Professional Communication in Electronic Media/Social Media
University of Duisburg-Essen
Duisburg, Germany
Email: bjoern.ross@uni-due.de

**Stefan Stieglitz**

Professional Communication in Electronic Media/Social Media
University of Duisburg-Essen
Duisburg, Germany
Email: stefan.stieglitz@uni-due.de

## Abstract

Information systems (IS) are widely used in organisations to improve business performance. The steady progression in improving technologies like artificial intelligence (AI) and the need of securing future success of organisations lead to new requirements for IS. This research in progress firstly introduces the term AI-based services (AIBS) describing AI as a component enriching IS aiming at collaborating with employees and assisting in the execution of work-related tasks. The study derives requirements from ten expert interviews to successful design AIBS following Design Science Research (DSR). For a successful deployment of AIBS in organisations the D&M IS Success Model will be considered to validated requirements within three major dimensions of quality: Information Quality, System Quality, and Service Quality. Amongst others, preliminary findings propose that AIBS must be preferably authentic. Further discussion and research on AIBS is forced, thus, providing first insights on the deployment of AIBS in organisations.

# 1   Introduction

Information systems (IS) have been used for years in large parts of organisations. "An information system can be seen as a system comprising human beings and/or machines which use and/or produce information" (Aram and Neumann 2015). When implemented in organisations, IS speed up business processes and thus save the adopting organisations a great amount of time (Neumann et al. 2014) and thereby money. In addition, benefits can be enhanced through the deployment of Artificial Intelligence (AI). There is no consensus on a precise definition, but AI is used for diverse objectives such as solving domain-independent problems, learning from the environment or interacting with other systems and humans (Dellermann et al. 2019). When used as components enriching IS in organisations, we relate to the expression AI-based services (AIBS), with the main objective of collaborating with employees and assisting in the execution of work-related tasks. AIBS assist employees in answering frequently asked questions by customers (Tsaih and Hsu 2018) which leads to a more effective and efficient way to handle inquiries and thus generates cost savings. AIBS are used for predicting desirable product preferences of customers (Nguyen and Sidorova 2018). Based on already existing information about customers, suitable related or new products can be offered. A targeted approach is possible, which is no longer determined by a subjective impression. AIBS are already used in the context of IT Service Management. Employed in the context of incident management, it massively accelerates the categorisation process (Frick et al. 2019), leading to time savings for employees.

Though AIBS generates numerous benefits for various areas in organisations focused on the collaboration between users and IS, no proper formulated guidelines for designing AIBS enriching IS in organisations can be found in the scientific literature. Securing the success in this fast-moving era, it is necessary to provide general guidelines precisely formulating which factors must be observed when planning and developing AIBS. To close this research gap, we propose suitable recommendations following the general guidelines for conducting Design Science Research (DSR). As contribution, results can be used by researchers as well as organisations and developers to design and implement AIBS. The development of our guidelines is based on qualitative research using ten face-to-face expert interviews as well as the review of existing literature. Theory-guided research recommends interpretation on two levels: results of one's own survey and conclusions of existing theories (Kohlbacher 2005). The measurement of IS effectiveness is a relevant component for both research and organisations. In this context, the DeLone & McLean IS Success Model (D&M IS Success Model) can be considered to measure the success of applied IS in organisations (DeLone and McLean 2003). To ascertain the success of AIBS in IS, the developed guidelines will be assigned to the proposed categories of the model and validated if the propositions fit the success metrics. In this context, three major dimensions of IS quality are considered: (1) Information Quality, (2) System Quality, and (3) Service Quality. To formalise the overall goal of this paper, we derived the following research question:

**RQ: What are the requirements that need to be considered to design AI-based services enriching information systems in organisations?**

The paper is structured as follows. First, we provide a theoretical overview on AI and AIBS as well as IS success in organisations. Second, we present the research design and the preliminary results including derived propositions. Last, we outline the next steps of our research process.

# 2   Theoretical Background

## 2.1   Artificial Intelligence and AI-based services

We define AIBS as components enriching information systems which use intelligent behaviour to assist people in fulfilling tasks. Recently, interest in AI has renewed as building systems has become more practical (Knijnenburg and Willemsen 2016) especially due to remarkable improvements in machine learning algorithms (Goodfellow et al. 2016). There is no generally accepted definition of the term AI since researchers from different disciplines have varying opinions and perceptions. As an omnipresent concept (Maedche et al. 2019), AI consists of multiple subfields which can be subdivided into various dimensions like thinking humanly, thinking rationally, acting humanly or acting rationally (Russel and Norvig 2016). AI can be used in various domains, most notably combining non-specialised and task-specialised intelligence as well as interacting with users and other (information) systems (Dellermann et al. 2019). The deployment of AI has become increasingly relevant for organisations to gain competitive advantages and maximise the market share. The growing amount of data and the possibility of gathering information in a short time (Nasirian et al. 2017), leads to the possibility of generating benefits which are highly relevant. Organisations exploiting the opportunities of AI will gain an advantage over competitors (Brynjolfsson and McAfee 2017). The development of new, enhanced services based on AI, generating benefits and simultaneously impacting collaboration within organisations (Seeber et al. 2018), enjoys an increasing attention. Oxford Dictionary (2019) defines service as the action of helping or doing work for someone. Adding AI capabilities to the services performed by information systems in organisations leads to the above definition of AIBS. The aim of using AIBS is for them to assist users in the execution of work-related tasks or even fulfil them entirely (Norman 2017). They are already applied in many organisations, and their impact has been validated. However, it has not yet been examined, how AIBS have to be properly designed to enrich IS in order to assist users, changing the nature of organisations (Seeber et al. 2018). Effectively integrating AIBS and its capabilities for collaboration, organisations find themselves at an ever-greater competitive advantage (Brynjolfsson and McAfee 2017).

## 2.2 Information Systems Success in Organisations

Organisations aim to improve business performance with IS. To this end, IS improve the effectiveness and efficiency of each organisation (Hevner et al. 2004) as well as facilitating collaboration by acting as communication and coordination systems (Aram and Neumann 2015). Thus, IS have three essential functions within an organisation: (1) Support the company's business operations, (2) Supporting managerial decision making, and (3) Support the achievement of strategic competitive advantages (Susanto and Meiryani 2019). Therefore, IS in organisations can be considered as business information systems (Aram and Neumann 2015). In this context, IS are composed of several information technologies (Orlikowski and Iacono 2001) which are applied to transmit, process or store information (Piccoli 2008). Considering the need for the ongoing improvement of IS to fulfil all business requirements, emphasises the impact of information technologies on business operations (Bjerknes et al. 1991).

In order to track the success of IS in organisations, it is necessary to conceptualise the quality of an IS within an organisation (DeLone and McLean 1992). To this end, the scholars developed the D&M IS Success Model "driven by a process understanding of information systems and their impacts" (DeLone and McLean 2002). Based on past research contributions, the D&M IS Success Model was updated. It describes three major dimensions of quality which each affect subsequent use and user satisfaction (DeLone and McLean 2003). Each quality dimension consists of various measurements. "Information quality" measures the overall content quality whereas "system quality" conceptualises the desired characteristics of an IS. Moreover, "service quality" explains the overall support delivered by the service provider. The dimension "use" measures the actual behaviour of a user whereas "user satisfaction" shows the opinion of the users about the applied IS. The dimension "net benefits" groups all impact measures, e.g. industry impacts, work group impacts, inter-organisational impacts or consumer impacts into one single category (DeLone and McLean 2003). To ensure that the developed design guidelines for AIBS follow the overall improvements of IS, results are assigned to the six dimensions of the D&M IS Success Model and it is validated whether they suit the success metrics.

## 3 Research Design

Despite the omnipresence of AIBS, there are still no clearly formulated guidelines for AIBS. To close this research gap, we produce design guidelines that can be used by researchers as well as organisations and developers to design and implement specific AIBS enriching IS in organisations focussing on the collaboration with employees. DSR advocates the construction of socio-technical artifacts (Gregor and Hevner 2013) to address relevant organisational problems (Hevner et al. 2004). Artifacts can be more specific (limited and less mature knowledge) and more abstract (complete and mature knowledge). Level 1 contains instantiations like software products or processes. Level 2 includes, for example, constructs and design principles. Level 3 describes design theories like mid-range and grand theory. The developed design guidelines, which contribute knowledge for operational principles or architecture, represent level 2. Following the DSR Methodology Process (Peffers et al. 2007), in this work in progress, we present the results of the first phase: identifying the problem. The development is based on qualitative research using ten face-to-face expert interviews as well as existing research results. Therefore, we reviewed relevant literature regarding topics of AI and adjoining subjects always concerning the enrichment of IS. Results are compared with the six dimensions of the D&M IS Success Model and validated if the developed requirements within the guidelines fit the various success metrics.

### 3.1 Expert Interviews

The method of the expert interviews has been established and grown in popularity as a valid method to obtain knowledge (Bogner et al. 2009). Especially in the exploratory phase of research, it is an efficient and concentrated method to collect relevant data (Bogner et al. 2009). The term expert describes someone who has an advantage of knowledge in the investigated field of research (Meuser and Nagel 2009). In this study, experts are individuals who have a special knowledge on their job, the company's structure, internal processes and, most important, where AIBS can be applied in IS to improve business performance. All interviews were conducted in face-to-face sessions and took place at the workplaces of the experts. This facilitates providing assistance to the interviewee and ensuring familiar conditions. The interviews took between 35 and 65 minutes. After conducting the interviews, they were transcribed and anonymised. Respecting data privacy protection, all audio recordings were deleted afterwards. The material will completely be coded in MAXQDA version 18.

### 3.1.1 Selection Process

Initially, eligibility criteria were first defined to select suitable companies. These companies should use a variety of internal IS to fulfil the daily work. It should be ensured that IS are already enriched by AIBS or future adoptions of AIBS are planned. Based on these factors, a large German retail holding organisation was selected which owns equity interests in further companies. Here, we chose companies focussing on various areas within the holding organisation: agricultural trade (C1), animal husbandry advisory (C2), consulting energy products (C3), animal feed advisory (C4), construction services (C5), wholesale e-commerce (C6) and agricultural machinery distribution (C7). To gain a holistic picture, we conducted ten interviews with experts working in management level with a minimum of three years of experience, because when the expert has a long tenure in a key position, "opportunities for expanding the researcher's access to the field may well also be unearthed in the interview" (Bogner et al. 2009). We acquired two project managers (E1/C2 [male, 28 years old, tenure of 8 years], E2/C1 [f, 25, 8]), three managing directors (E3/C3 [m, 35, 10], E7/C6 [m, 40, 21], E8/C1 [m, 43, 19]), three heads of divisions (E4/C4 [f, 40, 9], E9/C3 [f, 30, 4], E10/C5 [m, 43, 18]), and finally two managers (E5/C7 [m, 57, 5], E6/C7 [m, 47, 15]). Participants were 39 years old on average, with three female and seven male experts.

### 3.1.2 Semi-structured interviews

Semi-structured interviews are a flexible, accessible and intelligible as well as the most effective and efficient way to get relevant information from participants (Qu and Dumay 2011). The method should ensure that all relevant aspects were captured to generate comparable responses and ascertain that the following coding process is simplified. In preparation, interviews involve creating "questioning guided by identified themes in a consistent and systematic manner" (Qu and Dumay 2011). Therefore, a prefixed guide with central questions was developed, considering literature from the method of expert interviews and prior experiences of the researchers. The guide was divided into 9 parts: (1) Introduction of the interviewer and brief summary of the purpose of the research, as the participants had already received relevant information when they were recruited. (2) Self-introduction of the interviewee, including career development, current responsibilities in the company as well as demographic data. (3) Definition of AIBS and prior experience, followed by the authors' explanation of AIBS to ensure the same level of knowledge among all participants. (4) Areas in which AIBS are applied in organisations and which IS are enriched focusing on collaboration with employees. (5) Adoption and acceptance of AIBS and which problems might arise when enriching IS and collaborating with AIBS. (6) Advantages, disadvantages and dangers when collaborating with AIBS in IS. (7) How AIBS need to be developed in order to collaborate with them on a daily basis in IS. (8) Responsibility for an implementation and what an introduction looks like. (9) Conclusion of the interview: Possibility for the interviewee to ask further inquiries followed by a debriefing.

### 3.1.3 Coding

We used qualitative content analysis to evaluate the interviews as the most broad and exact way to analyse qualitatively collected material (Mayring 2015). This method orders the data according to certain empirically and theoretically reasonable points. The data is analysed using codes, which represent words or short phrases for attributes of language-based or visual data (Saldaña 2009) aiming at reducing the intricacy of vocabulary in the field and in the data by identifying one or multiple core categories (Flick 2013), finally leading to design guidelines. A initial list of general codes is created and collected within a codebook and maintained by one researcher as editor who is responsible for updating, revising and maintaining the list of codes within the group during the research process (Guest and MacQueen 2008). The coding is divided into two cycles: The first cycle takes place during the initial coding of the data. The second cycle focuses on pattern coding for categorisation of coded data. Following the codes-to-theory model (Saldaña 2009), we are currently in the first cycle carrying out initial coding. For interpreting what respondents mean, researchers need to have extensive knowledge in the subject matter (Campbell et al. 2013). Therefore, the authors have a strong background on collaboration systems, AI and AIBS as well as their utilisation in organisations. The coding is collaboratively done by two researchers. On the one hand, the effort for the coding process is distributed, on the other hand, different perspectives on the qualitative data are ensured.

# 4 Preliminary Results

## 4.1 Authenticity and Trust Perception

The first results show that the collaboration with an AIBS must preferably be authentic, since the method of communication differs between the individual colleagues. Besides, since AIBS are assisting users in their daily work, for example by generating recommendations, decisions or even fulfilling tasks autonomously, trust in such systems has to be as high as possible. One expert stated that *"[understanding the outcome is] very important! On the one hand, users can understand how the system came up with the decision, on the other hand, the users' level of knowledge is adjusted" (E1)*[1]. Authenticity and trust are interdependent and are critical for users to establish trust (Wünderlich and Paluch 2017). To positively affect authenticity and trust perception as well as use intention, the decision-making process has to be as transparent as possible (Wünderlich et al. 2013). As *system quality* of the D&M IS Success Model measures desired characteristics, authenticity and trust perception can be assigned to this dimension. We provide the following proposition as an approach to respond to our research question:

**P1: The collaboration with an AIBS enriching IS in organisations must be as authentic as possible to gain trust in such technologies.**

## 4.2 Safety, Security and Privacy Factors

A frequently mentioned topic in the interviews was the experts' concerns about safety, security and privacy factors. Talking about safety and security factors, the experts were less cautious. In the opinion of the experts, it was much more important to clarify which (personal) data is processed, where, how and by whom. On expert outlined *"so there will be a lot of scepticism, because everyone is afraid that personal data will be published. So everyone has quite a bit of respect" (E7)*. Within the D&M IS Success Model, *information quality* captures content issues of which safety, security and privacy factors are essential components. Being in control of personal data, its use and disclosure is mandatory (Cavoukian 2008). The interviewees indicated that adequate communication had to take place before the introduction of AIBS and that the legal basis had to be clarified in advance. We thus propose:

**P2: Data processed by AIBS enriching IS in organisations must be legally clarified and expounded to users.**

## 4.3 Enhanced Performance

Another prevalently mentioned point was the requirement that users should learn through the utilisation of AIBS. Experts underlined their statement by picturing situations in which users can better prepare for upcoming appointments and pay attention to matters they have previously disregarded. As one expert explained, *"I could well imagine [to learn from a system], so you get a well-grounded result" (E10)*. Since enhanced performance has a positive effect on employees and organisations, this attribute can be viewed as part of the concept of *net benefits* in the D&M IS Success Model. Norman (1994) already pointed out that technology is an instrument to enhance people's performances. Siddike et al. (2018) add that the interaction helps to boost performance at work. From this we derive following proposal:

**P3: Content presented by AIBS enriching IS in organisations must be presented appropriately to positively affect users' performances in their work.**

# 5 Next Steps of Research Process

In this research-in-progress paper, we presented the first phase of the DSR Methodology Process by identifying the problem and presented further research by giving insight on defining objectives of the solution. The analysis of the expert interviews is still at the beginning, yet this study encourages further discussion and research on AIBS. We are currently in the first cycle of the codes-to-theory model, conducting initial coding following the previously created codebook. In the next step, the analysis of the qualitative text data, including tests on reliability and validity, will be finalised. Based on the interviews, design guidelines are conducted representing level 2 artifacts of DSR contribution types. Findings will be compared and supported with existing research. Results will be linked with the six dimensions of the D&M IS Success Model to validate if the developed requirements within the guidelines fit in the various success metrics or whether there are new aspects that have not yet been considered. Propositions are assigned to the measures of the model. We will emphasise how the derived proposition supports the design of AIBS focussing on collaboration, in contrast with IS in general. However, further points must be considered when planning and developing AIBS enriching IS aiming at collaborating with employees, such as technical aspects, ethics and politics, user characteristics as well as requirements potentially exposed by the interviews. Overall, IS and organisations benefit when adapting AIBS. However, it has not yet been examined how AIBS must be designed to assist people fulfilling tasks. As a contribution, we aim at closing this research gap. Preliminary results already provide aspects considered when conducting AIBS in IS but do not provide a holistic picture. Research here offers promising results. The final guidelines will contain relevant requirements that need to be considered to design AIBS enriching IS in organisations with the main objective of collaborating with employees and assisting in the execution of work-related tasks.

---

[1] Excerpts from the German interviews have been translated into English for the reader's convenience.

# 6　References